\documentclass[10pt, a4paper]{article}

\usepackage{lrec-coling2024} % this is the new style [review]
\usepackage{amsfonts}
\usepackage{amsmath}
\usepackage{graphicx}
\usepackage{booktabs}
\usepackage{multirow}
\usepackage{xcolor}
\usepackage{algorithm}
\usepackage{algorithmic}

\title{Generative Multimodal Entity Linking}

\name{Senbao Shi, Zhenran Xu, Baotian Hu\textsuperscript{*}\thanks{*Corresponding author}, Min Zhang} 

\address{Harbin Institute of Technology, Shenzhen \\
         shisenbaohit@gmail.com, xuzhenran@stu.hit.edu.cn, \{hubaotian, zhangmin2021\}@hit.edu.cn\\}

\abstract{
Multimodal Entity Linking (MEL) is the task of mapping mentions with multimodal contexts to the referent entities from a knowledge base. 
Existing MEL methods mainly focus on designing complex multimodal interaction mechanisms and require fine-tuning all model parameters, which can be prohibitively costly and difficult to scale in the era of Large Language Models (LLMs). 
In this work, we propose GEMEL, a Generative Multimodal Entity Linking framework based on LLMs, which directly generates target entity names. 
We keep the vision and language model frozen and only train a feature mapper to enable cross-modality interactions. 
To adapt LLMs to the MEL task, we leverage the in-context learning capability of LLMs by retrieving multimodal instances as demonstrations. 
Extensive experiments show that, with only $\sim$0.3\% of the model parameters fine-tuned, GEMEL achieves state-of-the-art results on two well-established MEL datasets (7.7\% accuracy gains on WikiDiverse and 8.8\% accuracy gains on WikiMEL).
The performance gain stems from mitigating the popularity bias of LLM predictions and disambiguating less common entities effectively.
Further analysis verifies the generality and scalability of GEMEL.
Our framework is compatible with any off-the-shelf language model, paving the way towards an efficient and general solution for utilizing LLMs in the MEL task. Our code is available at \url{https://github.com/HITsz-TMG/GEMEL}.
 \\ \newline \Keywords{multimodal entity linking, large language models} }

\begin{document}

\maketitleabstract

\section{Introduction}

% EL

\textbf{Entity Linking} 
has attracted increasing attention in the natural language processing community, which aims at linking entity mentions in a document to referent entities in a knowledge base (KB)~\cite{EL}. 
It is a fundamental component in applications such as question answering~\cite{EL_for_QA}, relation extraction~\cite{EL_for_RE} and semantic search~\cite{EL_for_SS}.

\begin{figure}[!t]
\centering
\includegraphics[width=1\linewidth]{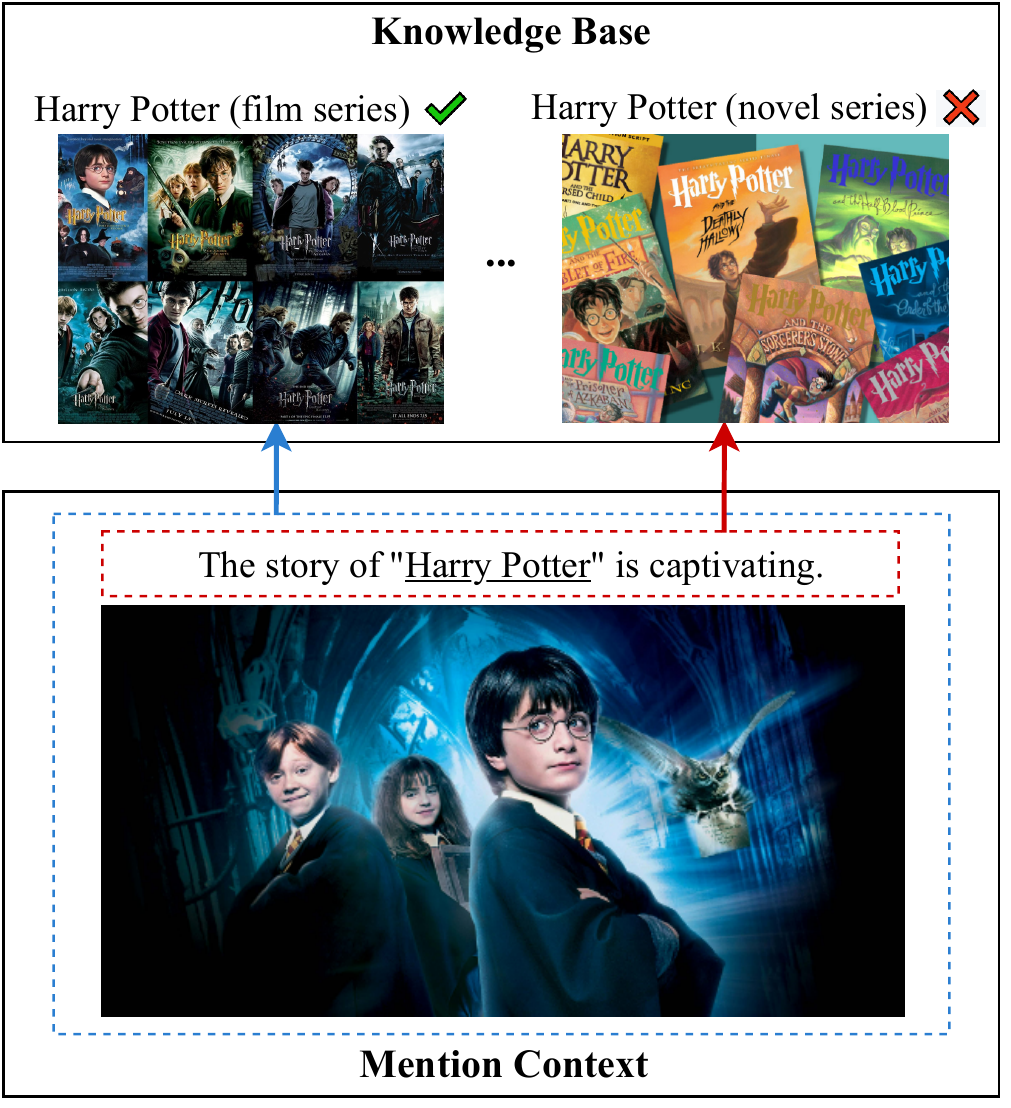}
 \caption{An example of multimodal entity linking with mention \underline{underlined} in the text. 
 Based solely on text, it is hard to determine whether “Harry Potter” should be linked to the films or the novels.
 With image as context, ``Harry Potter'' can be easily linked to the Harry Potter film series.} 
\label{fig:MEL_demo}
\end{figure}

Existing EL methods mainly focuses on textual modality and has been proven to be successful for well-formed texts~\cite{BLINK, GENRE}.
However, with the popularity of multimodal information on social media platforms,
more ambiguous mentions appear in the short or coarse text.
Due to the vast number of mentions arising from incomplete and inconsistent expressions, the conventional text-only EL methods cannot address cross-modal ambiguity, making it difficult to link these mentions accurately~\cite{JMEL}.
To address this issue, the task of \textbf{Multimodal Entity Linking} (MEL) has been proposed, which links mentions with multimodal contexts to their corresponding entities~\cite{DZMNED}. 
In Figure \ref{fig:MEL_demo}, based solely on textual modality, it is difficult to determine whether ``Harry Potter'' should be linked to the corresponding film series or the novel series. 
However, using multimodal information allows for correctly associating it with the film series.

Although some recent methods have achieved promising performance for the MEL task~\cite{DZMNED, JMEL, wang-etal-2022-wikidiverse, MEL-GHMFC-sigir}, challenges still exist. 
First, these methods often involve a two-stage process of candidate entity retrieval and re-ranking, 
which can be further improved in efficiency. 
For example, previous works~\cite{wang-etal-2022-wikidiverse, M3EL} use various sparse and dense retrievers and then combine candidate entities, followed by a re-ranking process.
Second, prior MEL methods~\cite{IMN, MEL-GHMFC-sigir} involve designing complex multimodal interaction mechanisms and training all model parameters, which is prohibitively costly and difficult to scale in the era of large language models (LLMs).

Recently, Vision-Language Models (VLMs) \cite{wang2022image,flamingo, ofa} trained on an enormous amount of image-text data have shown impressive results in various multimodal tasks. 
However, training a Large Vision-Language Models (LVLM) (e.g., Flamingo~\cite{flamingo}) from scratch is resource-intensive. 
To alleviate this issue, previous works~\cite{linear, magma, blip-2} propose that we can construct a LVLM based on the text-only LLM by transforming the visual information into the textual representation space of LLM. 
Through this approach, LLMs can effectively comprehend visual information and address multimodal tasks.
To step forward this direction, adapting LLMs to the MEL task and training models in a parameter-efficient manner have become a promising research direction.

Motivated by the analysis above, we propose a simple yet effective \textbf{Ge}nerative \textbf{M}ultimodal \textbf{E}ntity \textbf{L}inking framework (GEMEL) based on LLMs. 
Given the multimodal mention contexts, GEMEL can leverage the capabilities of LLMs from large-scale pre-training to directly generate corresponding entity names. 
We freeze the vision encoder and the language model, and only train a feature mapper to project visual features into a soft prompt for the LLM input.
Additionally, we utilize the in-context learning (ICL) capability by constructing multimodal demonstration examples to guide the LLMs to better comprehend the MEL task. 
Experimental results demonstrate that GEMEL can effectively integrate multimodal information based on LLMs to improve MEL performance, achieving state-of-the-art results on two well-established MEL datasets. 
Further studies reveal the popularity bias in entity predictions of LLMs, significantly under-performing on rare entities. 
GEMEL not only excels at common entity but also mitigates the popularity bias of LLMs to further boost performance on the MEL task.
Our framework is parameter-efficient and model-agnostic, and it can be transferred to any larger or stronger LLMs in the future.

In summary, the contributions of our work are as follows:

\begin{itemize}
    \item We propose GEMEL, a simple yet effective framework that utilizes a generative LLM to address the MEL task. To the best of our knowledge, this is the first work to introduce generative methods based on LLMs in the MEL task.
    \item Extensive experiments show that with only $\sim$0.3\% of the model parameters fine-tuned, GEMEL achieves state-of-the-art results on two well-established MEL datasets (7.7\% accuracy gains on WikiDiverse and 8.8\% accuracy gains on WikiMEL), exhibiting high parameter efficiency and strong scalability.
    \item Further studies reveal the popularity bias in LLM predictions, which our framework can effectively mitigate, thereby enhancing overall performance in the MEL task.
\end{itemize}

\begin{figure*}[t]
\centering
\includegraphics[width=\linewidth]{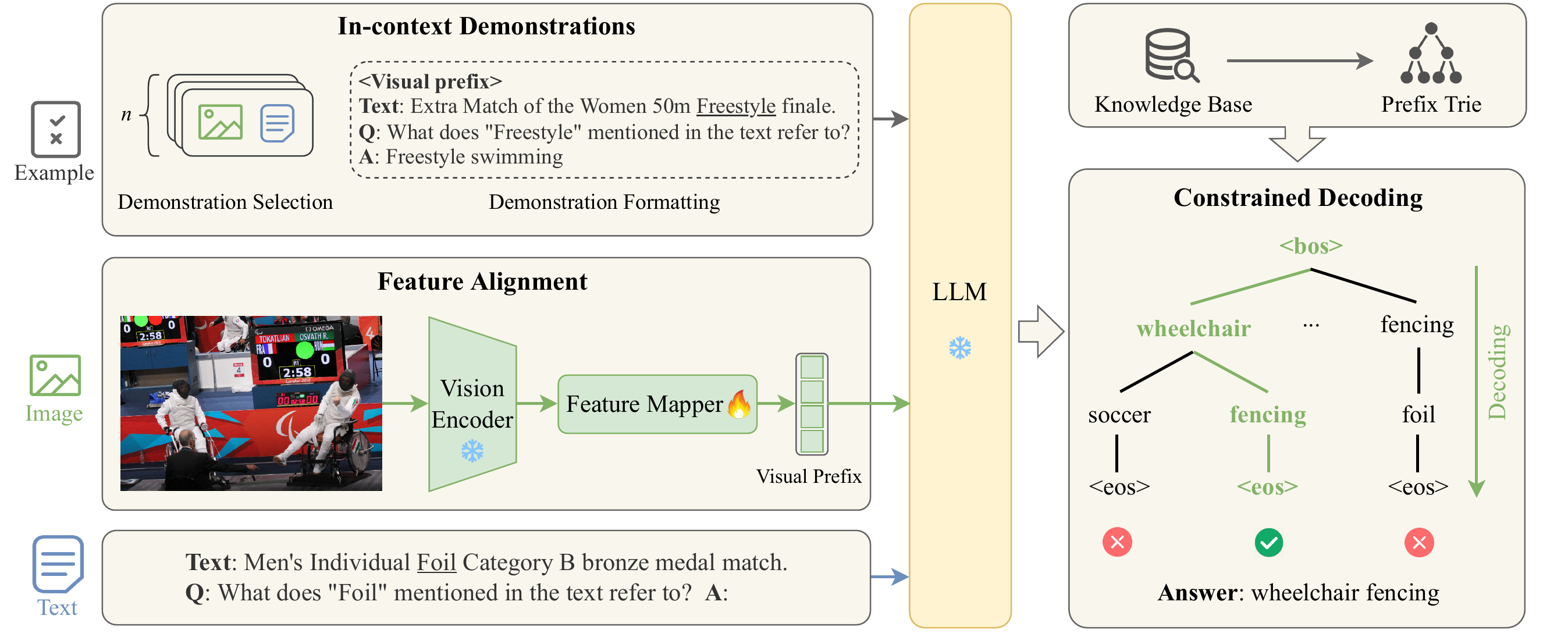}
 \caption{Overview of the GEMEL method. 
 Given the multimodal mention context, GEMEL first uses a feature mapper to transform image features to visual prefix in the textual space.
 Then, GEMEL leverages the capabilities of LLM to directly generate the target entity name (e.g. ``wheelchair fencing''), with $n$ retrieved multimodal instances as in-context demonstrations. 
GEMEL applies a constrained decoding strategy to efficiently search the valid entity space.
The mention in the text is \underline{underlined}.} 
\label{fig.GEMEL}
\end{figure*}

\section{Related Work}

\subsection{Textual EL}
Most previous EL methods follow a two-stage ``retrieve and re-rank'' pipeline~\cite{BLINK, extend, nlpcc, EPGEL}. 
The initial stage involves retrieving candidate entities from the vast number of entities in the knowledge base. 
Subsequently, in the second stage, these candidates undergo a re-ranking process to determine the final entity linking results. 
However, these methods heavily rely on retrieval results during re-ranking, leading to potential error accumulation and performance degradation. GENRE~\cite{GENRE} introduces a generative approach and uses constrained beam search to directly generate entity names. 
However, when dealing with short and coarse text, accurately linking entities becomes challenging if based solely on textual modality~\cite{JMEL}, thus motivating the task of Multimodal EL~\cite{DZMNED}.

\subsection{Multimodal EL}
Multimodal EL was initially proposed by Moon et al.~\cite{DZMNED} to address the ambiguous mentions in short social media posts. Although existing MEL methods have made significant progress~\cite{wang-etal-2022-wikidiverse, MEL-GHMFC-sigir, M3EL, IMN, MMEL}, they still have two limitations that need to be addressed: 
1) Prior MEL approaches utilize complex models with many interworking modules~\cite{wang-etal-2022-wikidiverse, MEL-GHMFC-sigir, M3EL}, which can be further improved in efficiency. 
For example, Wang et al.~\cite{wang-etal-2022-wikidiverse} combine candidate entity lists from multiple retrieval algorithms and then re-rank;
2) Existing MEL methods mainly focus on improving multimodal fusion based on co-attention mechanisms~\cite{IMN,MEL-GHMFC-sigir}, 
but they require training all model parameters, which becomes difficult to scale in the era of LLMs. 
To overcome these limitations, we propose a general end-to-end MEL framework that is parameter-efficient and easily scalable.

\subsection{LLMs for Vision-language Tasks}
In recent years, LLMs have demonstrated notable empirical achievements in language understanding~\cite{wei2022emergent}, generation~\cite{llm_generation}, and reasoning~\cite{llm_reason}. 
These successes have inspired recent research in vision-language tasks.
Since LLMs can only perceive text, bridging the gap between natural language and other modalities is necessary.

A feasible approach is to transform visual information into languages with the help of expert models (such as an image captioning model).
For example, VideoChat-Text~\cite{videochat} enriches the video descriptions with a speech recognition model.
However, though using expert models is straightforward, it may result in information loss.

Another effective approach is to introduce a learnable interface between the vision encoder and the LLM to connect information from different modalities.
For instance, BLIP-2~\cite{blip-2} utilizes a lightweight trainable Q-Former to bridge the modality gap and
achieves remarkable performance on various vision-language tasks.
Furthermore, some methods~\cite{linear, llava, pandagpt} use a projection-based interface to close the modality gap. 
For example, LLavA~\cite{llava} adopts a simple linear layer to project image features into textual embedding space.
Overall, previous works demonstrate that it is a potential research direction to utilize frozen LLMs for vision-language tasks.

\section{Methodology}

We introduce our \textbf{Ge}nerative \textbf{M}ultimodal \textbf{E}ntity \textbf{L}inking (GEMEL) framework. 
As shown in Figure~\ref{fig.GEMEL}, 
GEMEL takes the multimodal mention context and several in-context demonstrations as input,
and directly generates the target entity name.
We keep the parameters of the LLM and vision encoder frozen and only train a feature mapper to map image features into the textual space. 
In Section~\ref{sec:Formulation}, we first present problem formulation. Then, we describe the two components in GEMEL: feature alignment (Section~\ref{sec:Alignment}) and language model generation (Section~\ref{sec:Generation}).

\subsection{Problem Formulation}
\label{sec:Formulation}
Multimodal Entity Linking (MEL) is the task of mapping mentions with multimodal contexts to the corresponding entities in a knowledge base (KB). Formally, we define $\mathcal{E}$ as the entity set of the KB, which typically comprises millions of entities. Each mention $m$ is characterized by the visual context $V_m$ and textual context $T_m$. Here, $V_m$ represents the associated image for $m$, and $T_m$ represents the textual spans surrounding $m$. The task of MEL is to output mention-entity pairs: $\{(m_i, e_i)\}_{i \in [1,n_m] }$, where each corresponding entity $e_i$ is an entry in $\mathcal{E}$ and $n_m$ represents the number of mentions. Here we assume each mention has a valid gold entity in the KB, which is the \emph{in-KB} evaluation problem. We leave the out-of-KB prediction (i.e., $nil$ prediction) to future work.

\subsection{Feature Alignment}
\label{sec:Alignment}

Given the inherent incapacity of the LLMs to directly process multimodal information, it becomes imperative to undertake feature alignment procedures on visual data. In our approach for feature alignment, the initial step involves the extraction of image features through utilization of a pre-trained vision encoder. Subsequently, these image features undergo projection into the textual embedding space through the employment of a lightweight feature mapper. Upon this transformation, the resultant features are then introduced into the LLM as a visual prefix, thereby facilitating the LLM's capacity to process visual information.

\textbf{Vision Encoder.} To extract visual features from an input image $V_m$ corresponding to the mention $m$, we utilize a pretrained visual backbone model which produces visual embeddings $f_\phi(V_m) \in \mathbb{R}^{d_v}$, where $d_v$ represents the hidden state size of the vision encoder. The weights of vision encoder $\phi$ are kept frozen. 

\textbf{Feature Mapper.} To facilitate cross-modal alignment and fusion, we employ a feature mapper to project visual features into a soft prompt, i.e., a visual prefix for the LLM input. Specifically, we train a feature mapper $\boldsymbol{W} \in \mathbb{R}^{d_v \times kd_t}$ to project visual embeddings $f_\phi(V_m)$ into ${f_\phi(V_m)}^{T}\boldsymbol{W} \in \mathbb{R}^{kd_t}$. The result is then reshaped into a visual prefix, which is a sequence of $k$ embeddings $\boldsymbol{v} = \{v_1, v_2, ..., v_k\}$, where each embedding shares the same hidden dimensionality $d_t$ as the text embeddings produced by the LLM for input tokens.

\subsection{Language Model Generation}\label{sec:Generation}

In the context of language model generation, the visual prefix derived from the feature alignment module is concatenated with text embeddings and subsequently fed into the LLM. 
Moreover, we construct a few in-context demonstrations, which serve to enhance the LLM's understanding of the MEL task. Importantly, this process facilitates the generation of target entity names without necessitating any alterations to the LLM parameters.
At test time, we employ a constrained decoding strategy to efficiently search the valid entity space.

\textbf{In-context Learning.} To let the LLM better comprehend the MEL task, we leverage its in-context learning (ICL) ability~\cite{GPT-3, icl_survey}, and construct a prompting template with $n$ demonstration examples from the training set.
The demonstration formatting, depicted in Figure \ref{fig.GEMEL}, includes the image and textual context of mention $m$, a hand-crafted question (i.e., ``What does $m$ mentioned in the text refer to?'') and the entity name as the answer. 
For demonstration selection, we take several sparse and dense retrieval methods into consideration:

\begin{itemize}
    \item \textbf{Random selection} will randomly select in-context demonstrations from the training set for each mention.
    \item \textbf{BM25\footnote{We adopt the implementation from \url{https://github.com/dorianbrown/rank_bm25}.}} is one of the most advanced sparse retrieval methods. We use all the mentions in the training set as the corpus and retrieve demonstrations based on the mention. 
    \item \textbf{SimCSE\footnote{We utilize the supervised RoBERTa-large checkpoint from \url{https://github.com/princeton-nlp/SimCSE}.}}~\cite{simcse} is a dense retrieval method for semantic matching. For a pair of mentions, we take the cosine similarity of the mention embeddings as the relevance score.
    
\end{itemize}

 For $n$ demonstrations and a new instance $q$, we sequentially concatenate the visual prefix $\boldsymbol{v}$ and text embedding $\boldsymbol{t}$ to obtain the LLM input $\boldsymbol{x}$:

\begin{equation}
    \boldsymbol{x} = [\boldsymbol{v}_1; \boldsymbol{t}_1; ...; \boldsymbol{v}_n; \boldsymbol{t}_n; \boldsymbol{v}_q; \boldsymbol{t}_q]
\end{equation}

\textbf{Large Language Model.} GEMEL takes an autoregressive LLM $p_\theta$, which was originally trained with the maximum likelihood objective on text-only data, and keeps its parameters $\theta$ frozen. With the input embeddings $\boldsymbol{x}$ and $N$ tokens of the entity name as the target output $\boldsymbol{y} = \{y_1, y_2, ..., y_N\}$, we can express the teacher forcing training objective as follows:

\begin{equation}
    \mathcal{L}_{teacher} = - \sum_{i=1}^{N} \log {p}_{\theta}(y_i|\boldsymbol{x}; \boldsymbol{y}_{<i})
\end{equation}

\textbf{Constrained Decoding.}
During the testing phase, if the LLM is allowed to choose any word from its vocabulary at each decoding step, it may generate output strings that are not valid identifiers.
To address this issue, we exploit Constrained Beam Search~\cite{seq2seq, GENRE}, a tractable decoding strategy to efficiently search the valid entity space.
We define our constrain in terms of a prefix trie $\mathcal{T}$ where nodes are annotated with tokens from the LLM vocabulary.
For each node $t \in \mathcal{T}$, its children represent all the allowed continuations from the prefix defined traversing $\mathcal{T}$ from the root to t.
We tokenize all entity names in the knowledge base to construct a prefix tree, ensuring that the content generated by the LLM consists of valid entity names, thereby matching specific entities in the knowledge base.

\section{Experiments}

\begin{table}[t] \centering

  % \resizebox{1\linewidth}{!}{

    \begin{tabular}{lcc} % 
    \toprule
     &\textbf{WikiDiverse} &\textbf{WikiMEL} \\
    \midrule
    
    \# Image-text Pairs      &7824   &22136 \\
    \# Mentions              &16327  &25846 \\
    \# Text Length (avg.)    &10.2   &8.2   \\
    \# Mentions (avg.)       &2.1    &1.2   \\
    
  \bottomrule
\end{tabular}

% }
  \caption{Statistics of WikiDiverse and WikiMEL, two multimodal entity linking datasets.}
  \label{tab:datasets}
\end{table}

%============================Main Results==================================

\begin{table*}[ht!]\centering

% \resizebox{0.8\linewidth}{!}{
\begin{tabular}{cccc}\toprule
\multirow{2}{*}{\textbf{Modality}} &\multirow{2}{*}{\textbf{Model}} &\multicolumn{2}{c}{\textbf{Top-1 Accuracy (\%)}} \\
\cmidrule(lr){3-4}
& &WikiDiverse &WikiMEL \\
\cmidrule[\heavyrulewidth]{1-4}

\multirow{4}{*}{Text}
&BERT~\cite{BERT}       &69.6\textsuperscript{~~}   &31.7\textsuperscript{~~}   \\
&BLINK~\cite{BLINK}     &70.9\textsuperscript{~~}   &30.8\textsuperscript{~~}   \\
&GENRE~\cite{GENRE}     &78.0\textsuperscript{*}    &60.1\textsuperscript{*}   \\
&GPT-3.5-Turbo-0613     &72.7\textsuperscript{~~}   &73.8\textsuperscript{~~}   \\

\cmidrule{1-4}
\multirow{5}{*}{Text + Vision}
&JMEL~\cite{JMEL}       &38.4\textsuperscript{~~}   &31.3\textsuperscript{~~}   \\
&DZMNED~\cite{DZMNED}   &70.8\textsuperscript{~~}   &30.9\textsuperscript{~~}   \\
&GHMFC~\cite{MEL-GHMFC-sigir}   &62.7\textsuperscript{*}   &43.6\textsuperscript{~~}   \\
&LXMERT~\cite{wang-etal-2022-wikidiverse}  &78.6\textsuperscript{~~}   &-\textsuperscript{~}      \\
&MMEL~\cite{MMEL}    &-\textsuperscript{~~}      &71.5\textsuperscript{~~}   \\

\cmidrule{1-4}          
Text + Vision   &GEMEL (ours)      &\textbf{86.3}\textsuperscript{~~}   &\textbf{82.6}\textsuperscript{~~}  \\

\bottomrule
\end{tabular}
% }
\caption{Model performance on the test set. 
\textbf{Bold} denotes the best results. 
``*'' means our implementation with official Github repositories. 
``-'' means not reported in the cited paper.}

\label{tab:main_results}
\end{table*}

\subsection{Experimental Setup}

We elaborate the experimental protocols from the following four aspects: \emph{Datasets}, \emph{Baselines}, \emph{Evaluation Metric} and the \emph{Implementation Details.}

\textbf{Datasets.} To assess the capability of GEMEL on the MEL task, we conduct experiments on two MEL datasets, WikiDiverse~\cite{wang-etal-2022-wikidiverse} and WikiMEL~\cite{MEL-GHMFC-sigir}. 
WikiDiverse is a human-annotated MEL dataset with diversified contextual topics and entity types from Wikinews. 
WikiMEL is a large human-verified MEL dataset extracted from Wikidata and Wikipedia.
Both Wikidiverse and WikiMEL datasets have been splited into training, validation, and test sets, with ratios of 8:1:1 and 7:1:2, respectively,
and our experimental setting follows the partition.
The statistics of datasets are summarized in Table~\ref{tab:datasets}.

\textbf{Baselines.} Following previous works~\cite{MEL-GHMFC-sigir, MMEL, wang-etal-2022-wikidiverse}, we compare our method to recent state-of-the-art methods, categorized as follows: (1) text-only methods relying solely on textual features, and (2) text + vision methods utilizing both textual and visual features. Specifically, we consider:

\begin{itemize}
    \item \textbf{BERT} (text-only) \cite{BERT} is used as a text encoder to capture textual features and then calculate the relevance score.

    \item \textbf{BLINK} (text-only) \cite{BLINK} adopts a bi-encoder for candidate entity retrieval, followed by a cross-encoder for re-ranking.

    \item \textbf{GENRE} (text-only) \cite{GENRE} is the first system that retrieves entities by generating their names in an autoregressive fashion.
    % In this paper, we reproduced the experiment using the original configuration.

    \item \textbf{GPT-3.5-Turbo-0613} (text-only) is a powerful LLM developed by OpenAI. We utilize \texttt{GPT-3.5-Turbo-0613} (hereafter referred to as \texttt{GPT-3.5}) to generate entity names directly and employ the same ICL configuration as in our framework.

    \item \textbf{DZMNED} (text + vision) \cite{DZMNED} uses a concatenated multimodal attention mechanism to combine visual, textual, and character features of mentions and entities.

    \item \textbf{JMEL} (text + vision) \cite{JMEL} employs fully connected layers to map the visual and textual features into an implicit space.

    \item \textbf{GHMFC} (text + vision) \cite{MEL-GHMFC-sigir} uses gated hierarchical multimodal fusion and contrastive training to facilitate cross-modality interactions and reduce noise of each modality.

    \item \textbf{LXMERT} (text + vision) \cite{wang-etal-2022-wikidiverse} is a bi-encoder framework based on the multimodal encoder LXMERT~\cite{LXMERT} for calculating the relevance scores between mentions and entities.

    \item \textbf{MMEL} (text + vision) \cite{MMEL} proposes a joint learning framework to solve the multi-mention entity linking task in the multimodal scenario.

\end{itemize}

\textbf{Evaluation Metric.} Following previous works \cite{DZMNED,MEL-GHMFC-sigir, MMEL}, we use Top-1 accuracy as the evaluation metric.

\textbf{Implementation Details.} Our GEMEL framework is implemented with PyTorch~\cite{pytorch}.
We employ Llama-2-7B~\cite{LLaMa-2} and CLIP ViT-L/14 ~\cite{CLIP} as our default LLM and vision encoder, respectively, unless otherwise stated.
For feature mapping, we employ a linear layer~\cite{linear}, leaving the exploration of more complex feature mappers for future research.
We adopt SimCSE~\cite{simcse} to retrieve a total of $n = 16$ demonstration examples, sorted in ascending order based on their relevance to the mention. 
We search the visual prefix length $k$ among [1, 2, 4, 8] and find $k = 4$ performs the best based on the development set.
We set the beam size to 5 during inference. 
Our models are trained with a batch size of 1 and 16 steps of gradient accumulation for 5 epochs on a single A100 GPU. We utilize the AdamW optimizer \cite{AdamW} with a learning rate of 1e-6 and a warmup ratio of 10\%.

\subsection{Main Results}

Table \ref{tab:main_results} presents the model performances on two MEL datasets. 
According to the experimental results, we can see that: 
First, GEMEL surpasses all other approaches and achieves state-of-the-art performance on both MEL datasets, with a 7.7\% improvement on WikiDiverse (78.6\% $\rightarrow$ 86.3\%) and a 8.8\% improvement on WikiMEL (73.8\% $\rightarrow$ 82.6\%), showing the effectiveness of our framework. 
This indicates that, by fine-tuning a feature mapper ($\sim$0.3\% of model parameters), GEMEL enables the frozen LLM to comprehend visual information effectively and efficiently, and then leverage it to enhance MEL performance.
Second, methods based on LLMs (namely \texttt{GPT-3.5} and GEMEL) demonstrate powerful performance in both textual modality and multimodality. 
In the text modality, \texttt{GPT-3.5} can match or even exceed the performance of previous multimodal methods.
We reckon there are two main reasons for this: 
1) Textual modality still plays a dominant role in the MEL task, while visual modality primarily serves as supplementary information;
2) LLMs pretrained on large-scale datasets can capture extensive language patterns, context, and knowledge, leading to outstanding performance in common entity prediction (see Section~\ref{sec:Common} for details).

Table \ref{tab:ablation_study} shows the ablation study results. 
First, removing visual information significantly impairs GEMEL's performance, which indicates the importance of visual information 
when the texual information is short and insufficient (see cases in Section~\ref{sec:Cases}).
Second, eliminating the demonstrations in prompts leads to a performance decrease of 6.1\% and 7.4\% on WikiDiverse and WikiMEL respectively. 
This suggests that providing a few in-context demonstrations facilitates the LLM in recognizing and comprehending the MEL task (see Section~\ref{sec:demo} for details).

%============================Ablation Study==================================

\begin{table}[t]\centering
\resizebox{\linewidth}{!}{
\begin{tabular}{lcc}\toprule
\multirow{2}{*}{\textbf{Model}} &\multicolumn{2}{c}{\textbf{Top-1 Accuracy (\%)}} \\
\cmidrule(lr){2-3}
&WikiDiverse &WikiMEL \\
\cmidrule[\heavyrulewidth]{1-3}
GEMEL           &86.3    &82.6   \\           
\cmidrule{1-3}
w/o Visual Information        &79.5    &74.2   \\
w/o In-context Learning       &80.2    &75.2   \\
\bottomrule
\end{tabular}
}
\caption{Ablation results of GEMEL.}
\label{tab:ablation_study}
\end{table}

\section{Analysis}

%============================LLMs==================================

\begin{table}[t]\centering

\resizebox{\linewidth}{!}{
\begin{tabular}{lccc}\toprule

\multirow{2}{*}{\textbf{LLM}} &\multirow{2}{*}{\textbf{Parameters}} &\multicolumn{2}{c}{\textbf{Top-1 Accuracy (\%)}} \\
\cmidrule(lr){3-4}
& &WikiDiverse &WikiMEL \\
\cmidrule[\heavyrulewidth]{1-4}
OPT        &6.7B   &82.7   &75.5   \\
Llama      &7B     &85.8   &82.2   \\
Llama-2    &7B     &86.3   &82.6   \\
\bottomrule
\end{tabular}
}
\caption{Results of different LLMs.} 
\label{tab:LLMs}
\end{table}

In this section, we will analyze the performance of our framework from four aspects. First, we will explore the generality and scalability of our framework. Second, we will investigate the influence of different demonstration selection methods in our framework. Subsequently, we identify the popularity bias in LLM predictions (i.e., significantly under-performing in tail entities), which our framework effectively mitigates. Finally, we  conduct case study and limitation analysis.

\subsection{Generality and Scalability}

To test the generality of GEMEL across different LLMs, we conduct experiments on various LLMs including OPT~\cite{OPT}, Llama~\cite{LLaMa}, and Llama-2~\cite{LLaMa-2}.
As shown in Table~\ref{tab:LLMs}, our GEMEL framework is generally effective for the widely-used LLMs.

%============================Vision Encoder==================================

\begin{table}[t]\centering

% \resizebox{1\linewidth}{!}{
\begin{tabular}{lcc}\toprule

\multirow{2}{*}{\textbf{Method}} &\multicolumn{2}{c}{\textbf{Top-1 Accuracy (\%)}} \\
\cmidrule(lr){2-3}
 &WikiDiverse &WikiMEL \\
\cmidrule[\heavyrulewidth]{1-3}
ResNet-101          &85.3   &82.1   \\
BEIT-large          &84.7   &81.0   \\
CLIP ViT-L/14       &86.3   &82.6   \\
\bottomrule
\end{tabular}
% }
\caption{Results of different vision encoders.} 
\label{tab:vision_encoder}
\end{table}

\begin{figure}[t]
\centering
\includegraphics[width=0.95\linewidth]{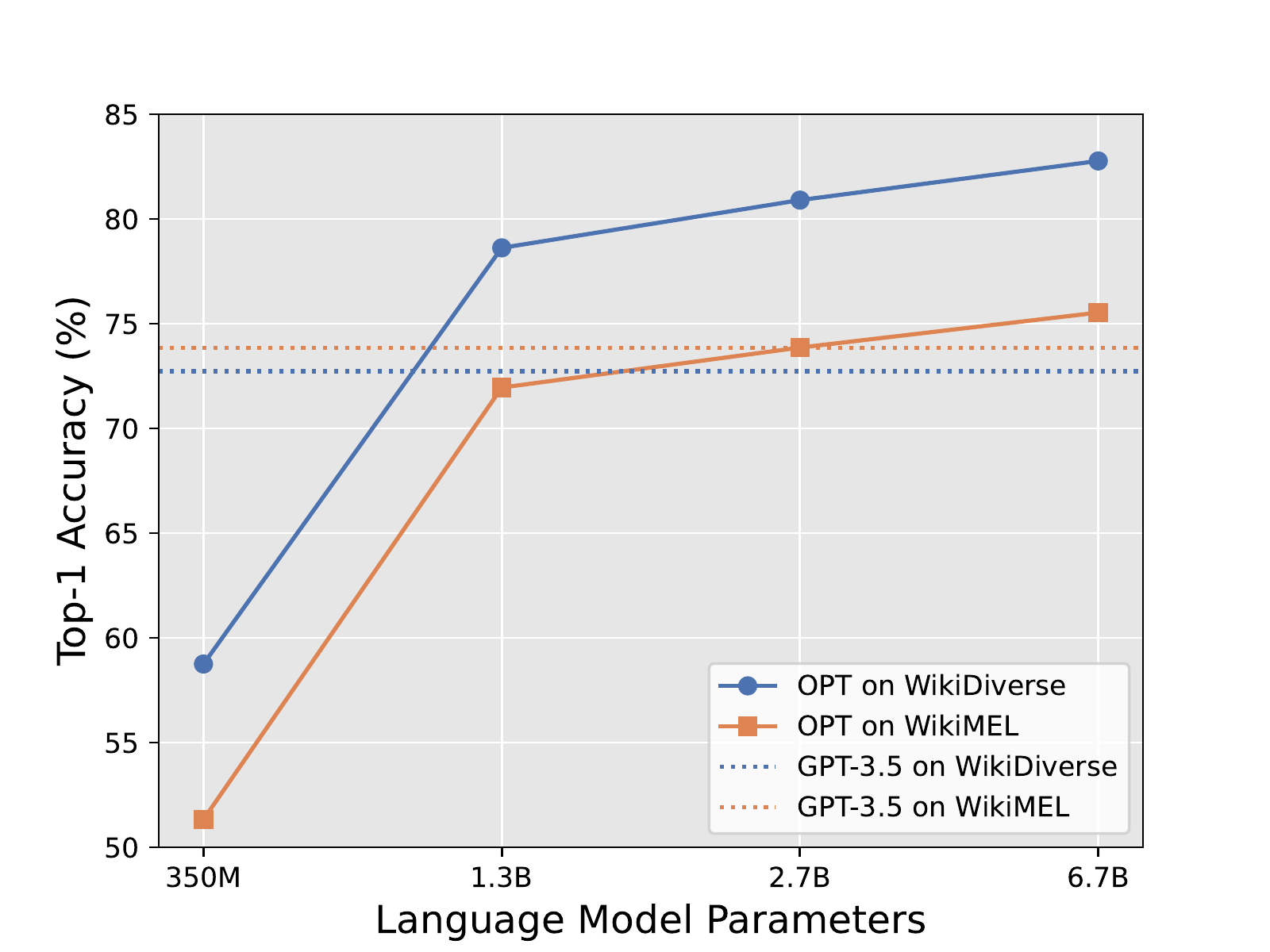}
 \caption{Results of scaling up parameters of OPT. As language models continue to scale up, GEMEL consistently exhibits enhanced performance in the MEL task.}
\label{fig:model_scale}
\end{figure}

%============================Demostration Selection==================================

\begin{table}[t]\centering
% \resizebox{1\linewidth}{!}{
\begin{tabular}{lcc}\toprule
\multirow{2}{*}{\textbf{Method}} &\multicolumn{2}{c}{\textbf{Top-1 Accuracy (\%)}} \\
\cmidrule(lr){2-3}
&WikiDiverse &WikiMEL \\
\cmidrule[\heavyrulewidth]{1-3}
No-Retrieval            &80.2    &75.2   \\
\cmidrule{1-3}    
Random Selection        &83.1    &78.3   \\           
BM25                    &85.3    &82.5   \\
SimCSE               &\textbf{86.3}    &\textbf{82.6}   \\
\bottomrule
\end{tabular}
% }
\caption{Results of different demonstration selection methods. \textbf{Bold} indicates the best performance.}
\label{tab:demostration_selection}
\end{table}

Different vision encoders may affect the model performance. We compare three widely-used types of vision encoders, ResNet-101~\cite{resnet}, BEiT-large~\cite{beit}, and CLIP ViT-L/14 image encoder~\cite{CLIP}.
As shown in Table~\ref{tab:vision_encoder}, our framework performs well with various visual encoders, with CLIP achieving the best results. Therefore, we use CLIP by default in our framework.

\begin{table*}[t]\centering

% \resizebox{0.85\linewidth}{!}{
\begin{tabular}{lcccc}\toprule
\multirow{2}{*}{\textbf{Model}} &\multicolumn{2}{c}{\textbf{WikiDiverse}} &\multicolumn{2}{c}{\textbf{WikiMEL}} \\
\cmidrule(lr){2-3}
\cmidrule(lr){4-5}
&Common Entity &Tail Entity &Common Entity &Tail Entity \\
\cmidrule[\heavyrulewidth]{1-5}

GPT-3.5-Turbo-0613                   &72.0   &37.5   &74.4   &65.5 \\
GEMEL (w/o Visual Information)                       &80.8   &40.0   &74.4   &68.8 \\
GEMEL                                &87.4   &57.5   &82.7   &75.2 \\

\bottomrule
\end{tabular}
% }
\caption{Accuracy (\%) of LLM-based methods on common entity and tail entity.
% The LLM-based methods display noticeable bias in the accuracy of predicting common entity and tail entity.
} 

\label{tab:tail_entity}
\end{table*}

As shown in Figure \ref{fig:model_scale}, we employ OPT models of varying scales (i.e., 350M, 1.3B, 2.7B, 6.7B) to assess the influence of language model scale on GEMEL performance. 
We do not perform the ICL prompting for the OPT 350M model due to its insufficient scale for demonstrating the ICL ability. 
It is evident that with the language model scaling up, GEMEL consistently improves its performance on the MEL task. 
Our framework even outperforms \texttt{GPT-3.5} on small-scale language models, as demonstrated by the results of OPT-1.3B on the WikiDiverse dataset.
This indicates that our framework is highly effective and model-agnostic, meaning it can be applied to larger or more powerful LLMs that may be released in the future, thereby further enhancing performance.
Given resource constraints, our investigation is limited to models up to 6.7B in scale, leaving the exploration of larger language models for future work.

\subsection{Demonstration Selection}\label{sec:demo}

To investigate the influence of demonstration selection on the performance of GEMEL, we conduct experiments using several sparse and dense retrieval methods: random selection, BM25, and SimCSE~\cite{simcse}. The results of demonstration selection are shown in Table~\ref{tab:demostration_selection}, where we also present the result of no-retrieval method (i.e.,~without ICL). The experimental findings indicate the following: 
1) Whether employing random selection, BM25, or SimCSE, all of these methods surpass the approach of no-retrieval. This indicates that incorporating in-context demonstrations facilitates LLMs in recognizing the format of the MEL task. 
2) Methods utilizing similarity retrieval outperform random selection. 
We believe that demonstrations retrieved based on similarity are more likely to include similar mentions and entity candidates, thus learning through similar demonstrations and enhancing MEL performance;
3) The retrieval approach utilizing SimCSE outperforms the BM25 method. 
We reckon that employing dense retrieval methods, such as SimCSE, enhances the comprehension of semantic meaning in mentions compared to sparse methods like BM25. 
This results in the retrieval of more relevant and higher quality demonstrations.

\begin{figure*}[t]
\centering
\includegraphics[width=\textwidth]{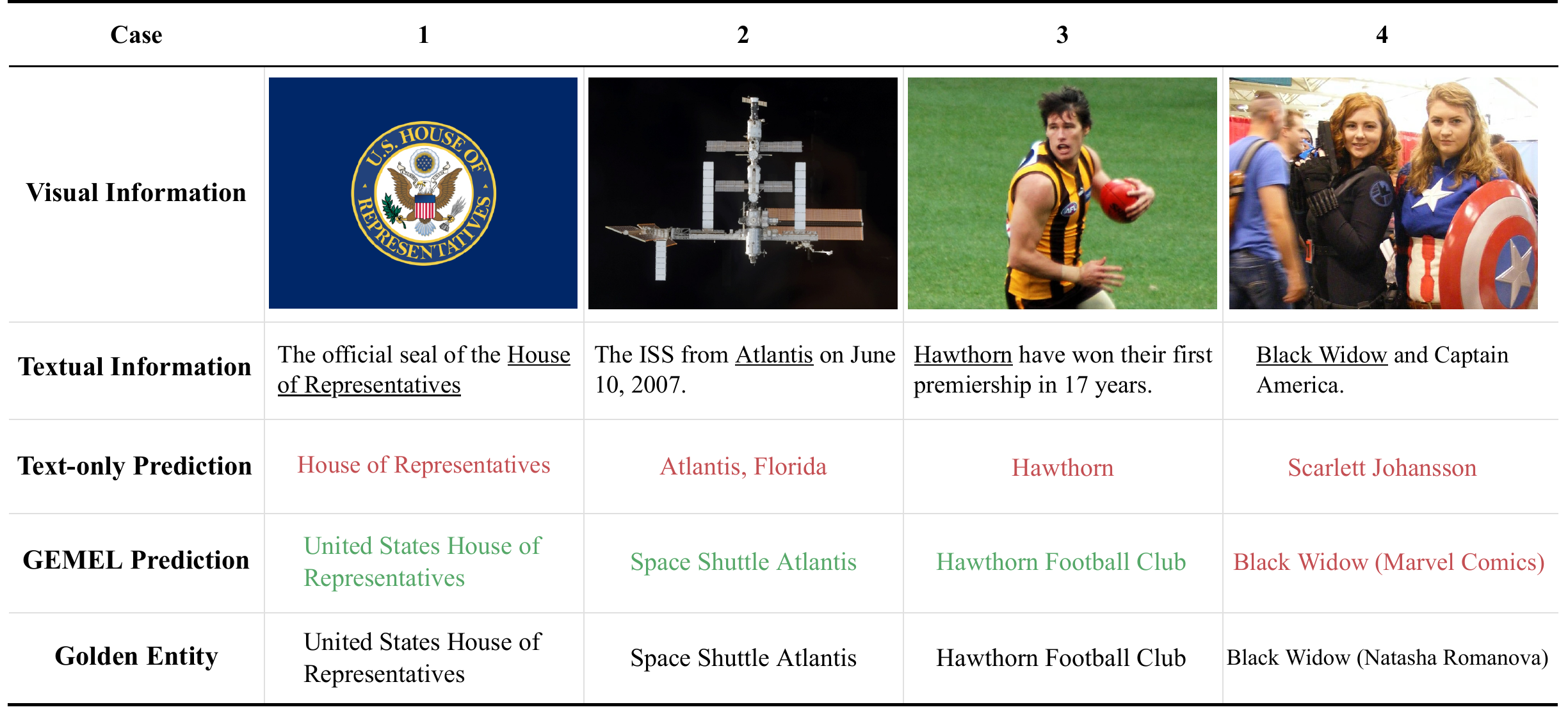}
 \caption{Case study. For \underline{underlined} mentions, \textcolor[HTML]{55A868}{green} and \textcolor[HTML]{C44E52}{red} text mean successful and failed predictions, respectively. The text-only prediction results are obtained by GEMEL without utilizing visual information.} 
\label{fig:case_study}
\end{figure*}

\subsection{Popularity Bias of LLMs}\label{sec:Common}

To explore why methods based on LLMs can exhibit remarkable performance in both textual modality and multi-modality, we conduct a statistical analysis of prediction outcomes using LLM-based approaches. 
Following previous work~\cite{tail_entity}, we tallied the occurrences of each entity in Wikipedia\footnote{We adopt the result from GENRE~\cite{GENRE} repository: \url{https://dl.fbaipublicfiles.com/GENRE/mention2wikidataID_with_titles_label_alias_redirect.pkl}.}.
Entities that appear less than 5\% of the total count are considered as \emph{tail entities}, while the rest are regarded as \emph{common entities}.
We conduct statistical analysis on the accuracy of \texttt{GPT-3.5}, GEMEL (w/o Visual Information), and GEMEL in terms of common entity and tail entity predictions. The results are shown in Table \ref{tab:tail_entity}.

Based on the statistical results, we can observe the following:
1) The LLM-based approaches demonstrate impressive performance in the textual modality, primarily attributed to LLM's exceptional performance in predicting common entity after large-scale pre-training. For instance, both \texttt{GPT-3.5} and GEMEL achieve accuracy rates exceeding 70\% in predicting common entity;
2) In terms of predicting tail entity, methods based on LLMs exhibit a popularity bias, i.e., significantly under-performing on less common entities. Taking \texttt{GPT-3.5}'s result on WikiDiverse as an example,  the prediction accuracy for common entity reaches 72.0\%, whereas for tail entity, it is only 37.5\%;
3) Our framework not only ensures significant performance in predicting common entities but also demonstrates substantial improvements in tail entity prediction. For instance, on the WikiDiverse dataset, compared to \texttt{GPT-3.5}, GEMEL exhibits notable enhancement in common entity prediction and a significant increase of 20.0\% (37.5\% $\rightarrow$ 57.5\%) in tail entity prediction. This indicates the effectiveness of our multimodal framework in mitigating the bias of LLMs towards tail entity prediction, thereby enhancing overall performance in the MEL task.

\subsection{Case Study}\label{sec:Cases}

In Figure \ref{fig:case_study}, we compare prediction results of GEMEL in textual modality and multimodality. 

Cases 1, 2, and 3 are examples where predictions are incorrect in the text-only modality but correct in the multimodality. 
It can be observed that when the information provided by the text is insufficient, visual modality can complement information and help eliminate ambiguity. 
Taking Case 1 as an example, in situations with only textual modality, the model struggles to determine the region to which ``House of Representative'' belongs. 
However, when visual information is introduced, GEMEL effectively utilizes visual information to accurately link the mention to a specific entity.

Case 4 predicts incorrectly in both textual modality and multimodality. It requires linking to the specific role of ``Black Widow'' based on multimodal information, which necessitates making judgments considering details such as attire.
This inspires us to explore more fine-grained multimodal information in future research.

\section{Conclusion and Future Work}

We propose GEMEL, a simple yet effective generative multimodal entity linking framework based on LLMs, which leverages the capabilities of LLMs to directly generate target entity names. 
Experimental results demonstrate that GEMEL outperforms state-of-the-art methods on two MEL datasets, exhibiting high parameter efficiency and strong scalability. 
Further studies reveal the existence of bias in LLMs predictions for tail entity, which our framework can effectively mitigate, thereby enhancing overall performance in the MEL task.
Moreover, our framework is model-agnostic, enabling its application to larger or more powerful LLMs in the future. Further research can explore how to mitigate bias in tail entity prediction for LLMs 
and extend GEMEL to more modalities (such as video, speech, etc.).

\section*{Acknowledgments}
We would like to thank Yunxin Li, Ziyang Chen, Qian Yang, Xinshuo Hu, and Yulin Chen for the valuable comments and useful suggestions.  
This work is supported by grants: Natural Science Foundation of China (No. 62376067).

\nocite{*}
\section*{References}\label{sec:reference}

\bibliographystyle{lrec-coling2024-natbib}
\bibliography{languageresource}

\end{document}